\documentclass[10pt,twocolumn,twoside]{IEEEtran}
\usepackage{authblk,amsmath,graphicx}
\usepackage{amssymb}
\usepackage{amsthm}
\usepackage[hyphens]{url}
\usepackage{caption}
\usepackage{float}
\usepackage{amsmath,graphicx,epsfig}

\newcommand{\R}{\mathbb{R}}
\DeclareMathOperator{\diag}{diag}
\newcommand{\bbm}{\begin{bmatrix}}
\newcommand{\ebm}{\end{bmatrix}}
\newtheorem{thm}{Theorem}[section]
\newtheorem{lemma}[thm]{Lemma}

\begin{document}
\title{A convex model for non-negative matrix factorization and dimensionality reduction on physical space}
\author{Ernie~Esser, Michael~M\"{o}ller, Stanley~Osher, Guillermo~Sapiro, Jack~Xin
\thanks{EE and JX were supported by NSF grants \mbox{DMS-0911277}, \mbox{PRISM-0948247}, MM by the German Academic Exchange Service (DAAD), SO and MM by NSF grants \mbox{DMS-0835863}, \mbox{DMS-0914561}, \mbox{DMS-0914856} and ONR grant \mbox{N00014-08-1119},  and GS was supported by NSF, NGA, ONR, ARO, DARPA, and \mbox{NSSEFF.}}%
\thanks{E. Esser and J. Xin are with the Department of Mathematics, University of California Irvine, 340 Rowland Hall, Irvine, CA 92697-3875. M. M\"{o}ller and S. Osher are with the Department of Mathematics, University of California Los Angeles, Box 951555, Los Angeles, CA 90095-1555. G. Sapiro is with the Department of Electrical and Computer Engineering, University of Minnesota, 200 Union Street SE, 4-174 Keller Hall, Minneapolis, MN 55455-0170. }}

\maketitle
\begin{abstract}
A collaborative convex framework for factoring a data matrix $X$ into
a non-negative product $AS$, with a sparse coefficient matrix
$S$, is proposed. We restrict the columns of the dictionary matrix $A$ to coincide
with certain columns of the data matrix $X$, thereby guaranteeing a physically
meaningful dictionary and dimensionality reduction.  We use $l_{1,\infty}$ regularization to select the dictionary from the data and show this leads to an exact convex relaxation of $l_0$ in the case of distinct noise free data.  We also show how to relax the restriction-to-$X$ constraint by initializing an alternating minimization approach with the solution of the convex model, obtaining a dictionary close to but not necessarily in $X$.  We focus on applications of the proposed framework to hyperspectral endmember and abundances identification and also show an application to blind source separation of NMR data.
\end{abstract}

\section{Introduction}
\label{sec:intro}
\IEEEPARstart{D}{imensionality} reduction has been widely studied in the signal processing and computational learning communities. One of the major drawbacks of virtually all popular approaches for dimensionality reduction is the lack of physical meaning in the reduced dimension space. This significantly reduces the applicability of such methods. In this work we present a framework for dimensionality reduction, based on matrix factorization and sparsity theory, that uses the data itself (or small variations from it) for the low dimensional representation, thereby guaranteeing the physical fidelity.  We propose a new convex method to factor a non-negative data matrix $X$ into a product $AS$, for which $S$ is non-negative and sparse and the columns of $A$ coincide with columns from the data matrix $X$.

The organization of this paper is as follows.  In the remainder of the introduction, we further explain the problem, summarize our approach and discuss applications and related work.  In Section \ref{sec:method} we present our proposed convex model for {\it endmember} (dictionary) computation that uses $l_{1,\infty}$ regularization to select as endmembers a sparse subset of columns of $X$, such that sparse non-negative linear combinations of them are capable of representing all other columns.  Section \ref{sec:l0} shows that in the case of distinct noise free data, $l_{1,\infty}$ regularization is an exact relaxation of the ideal row-$0$ norm (number of non-zero rows), and furthermore proves the stability of our method in the noisy case.  Section \ref{sec:results} presents numerical results for both synthetic and real hyperspectral data. In Section \ref{sec:extended} we present an extension of our convex endmember detection model that is better able to handle outliers in the data. We discuss its numerical optimization, compare its performance to the basic model and also demonstrate its application to a blind source separation (BSS) problem based on NMR spectroscopy data.

\subsection{Summary of the problem and geometric interpretation}
The underlying general problem of representing $X \approx AS$ with $A,S \geq0$ is known as non-negative matrix factorization (NMF).  Variational models for solving NMF problems are typically non-convex and are solved by estimating $A$ and $S$ alternatingly. Although variants of alternating minimization methods for NMF often produce good results in practice, they are not guaranteed to converge to a global minimum.

\begin{figure}[H]
  \centering
  \centerline{\includegraphics[width=9cm]{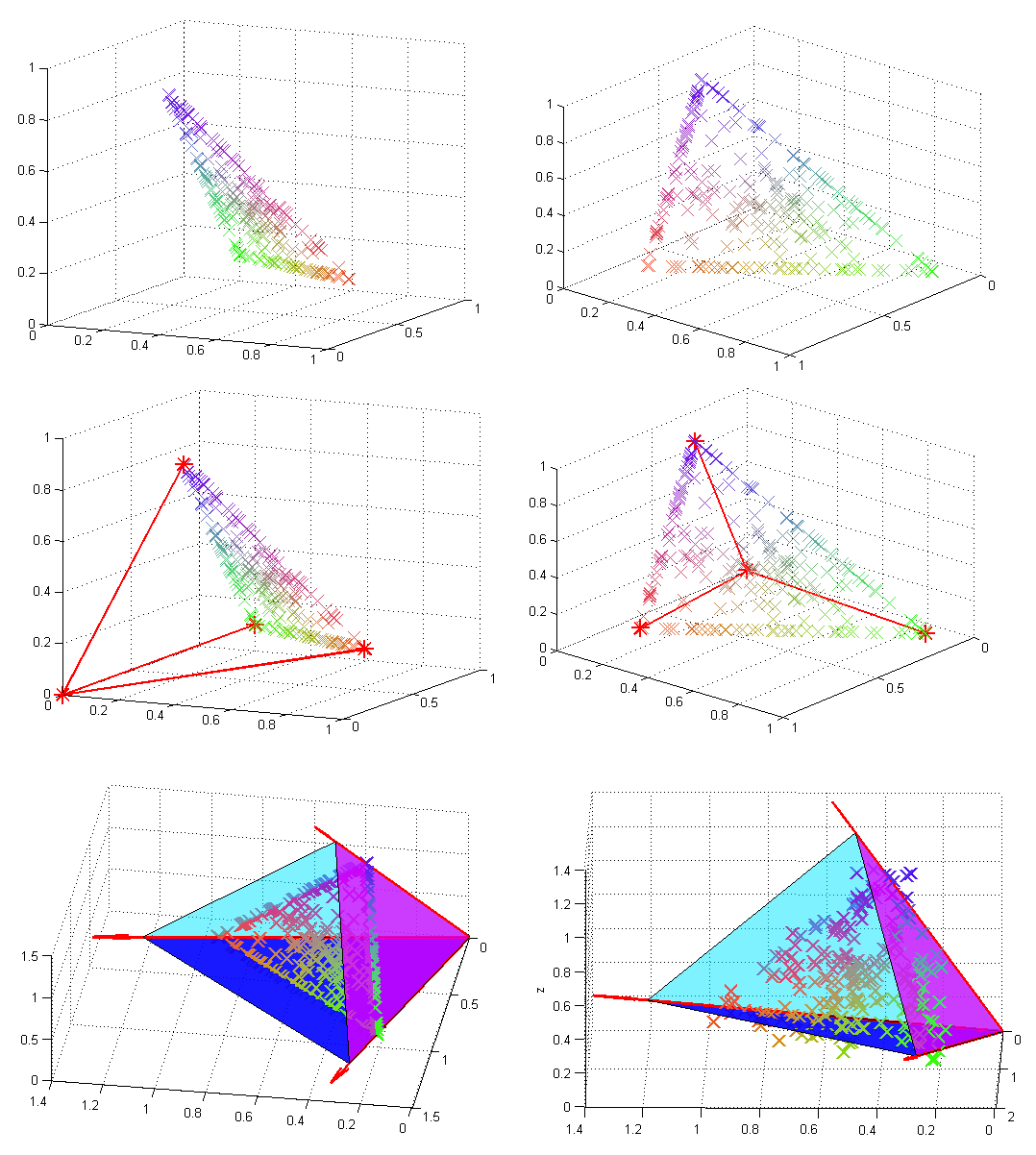}}
  \caption{Geometric interpretation of the endmember detection problem. First row: two different viewpoints for a data set in three dimensions, second row: same data set with the vectors that can express any data point as a non-negative linear combination in red, third row left: cone spanned by the three vectors containing all data, third row right: illustration of the cone with Gaussian noise added to the data, in this case not all points lie inside the cone anymore}
  \label{fig:geom}
\end{figure}

The problem can be greatly simplified by assuming a partial orthogonality condition on the matrix $S$ as is done in \cite{NN,winter99}. More precisely, the assumption is that for each row $i$ of $S$ there exists some column $j$ such that $S_{i,j} > 0$ and $S_{k,j} = 0$ for $k \neq i$.  Under this assumption, NMF has a simple geometric interpretation.  Not only should the columns of $A$ appear in the data $X$ up to scaling, but the remaining data should be expressible as non-negative linear combinations of these columns.  Therefore the problem of finding $A$ is to find columns in $X$, preferably as few as possible, that span a cone containing the rest of the data $X$.  Figure \ref{fig:geom} illustrates the geometry in three dimensions.

The problem we actually want to solve is more difficult than NMF in a couple respects.  One reason is the need to deal with noisy data.  Whereas NMF by itself is a difficult problem already, the identification of the vectors becomes even more difficult if the data $X$ contains noise and we need to find a low dimensional cone that contains most of the data (see lower right image in Figure \ref{fig:geom}). Notice that in the noisy case, finding vectors such that all data is contained in the cone they span would lead to a drastic overestimation of the number of vectors. Arbitrarily small noise at a single data point could already lead to including this vector into the set of cone spanning vectors. Thus, the problem is ill-posed and regularization is needed to handle noise.  In addition to small noise there could also be outliers in the data, namely columns of $X$ that are not close to being well represented as a non-negative linear combination of other columns, but that we do not wish to include in $A$.  Such outliers could arise from bad sensor measurements, non-physical artifacts or any features that for some reason we are not interested in including in our dictionary $A$.  Another complication that requires additional modeling is that for the applications we consider, the matrix $S$ should also be sparse, which means we want the data to be represented as sparse non-negative linear combinations of the columns of $A$.

\subsection{Our proposed approach}
We obtain the $X=AS$ factorization by formulating the problem as finding a sparse non-negative $T$ such that $X\approx XT$ and as many rows of $T$ as possible are entirely zero.  We want to encourage this so called \textit{row sparsity} of $T$ in order to select as few as possible data points as dictionary atoms.  We do this by using $l_{1,\infty}$ regularization.  This type of regularization couples the elements in each row of $T$ and is based on the recent ideas of collaborative sparsity (see for example \cite{Bach09} and the references therein). The particular $l_{1,\infty}$ regularization has been studied by Tropp in \cite{tropp06}, however, without considering non-negativity constraints and also not in the setting of finding a $T$ such that $X\approx XT$ for physically meaningful dimensionality reduction.  A strong motivation for using $l_{1,\infty}$ regularization instead of other row sparsity regularizers like $l_{1,2}$ is that in the case of distinct, noise-free data, the $l_{1,\infty}$ model is an exact relaxation of minimizing the number of non-zero rows in $T$ such that $X=XT$.  This exact relaxation is independent of the coherence of the columns of $X$.  Without the non-negativity constraint, a low coherence is crucial as is shown in \cite{tropp06}. The general setting $X\approx XT$ was proposed by Lin et al. in \cite{Lin10} for low rank approximation with the nuclear norm. However, the nuclear norm does not lead to row sparsity, and thus there is no obvious way to extract dictionary atoms from the minimizer.  Both the nuclear norm and $l_{1,\infty}$ approaches are addressing related but different problems.  Our main contribution is to apply the joint sparsity regularizer $l_{1,\infty}$ to the non-negative factorization setting $X\approx XT$ and thereby implicitly select certain columns of $X$ for the description of all columns of $X$.  We pose this as a convex optimization problem in $T$.  For practical reasons, we will need to perform some preliminary data reduction, explained in Section \ref{sec:method}, before carrying out the convex minimization.  The main idea, however, is to minimize over $T \geq 0$ the $l_{1,\infty}$ norm of $T$ plus some convex penalty on $X-XT$.  In the simplest case, we penalize $\|X-XT\|_F^2$.  We also propose an advanced noise model to handle the case where $X$ contains outliers.  Both models also incorporate a weighted $l_1$ penalty to encourage a sparser $T$ so that from the few columns of $X$ selected to represent the whole data, only a few are used per sample.

\subsection{Applications and related work}
Although we concentrate on hyperspectral imaging (HSI) and briefly discuss an application to blind source separation (BSS), our method is applicable in numerous areas, from biology to sensor networks.

For instance, one approach to text mining, that is the technique of extracting important information from large text data sets, is to reduce the number of relevant text documents by clustering them into content dependent categories using non-negative matrix factorization (see \cite{Xu03} for details). Our approach could potentially be used to not only cluster the large amount of documents by a similar factorization, but due to the dictionary being a part of the data, it would furthermore lead to a correspondence of every atom in the dictionary to a certain document. This correspondence might help a human analyzer judge the importance of each cluster. Therefore the physical meaning of the dictionary atoms would have an immediate advantage for the further analysis.

As another example, in \cite{Holzapfel08} non-negative matrix factorization is applied to spectrograms of many different musical sounds in order to obtain a spectral basis for musical genre classification. Again the physical fidelity of our approach could be interesting since it would provide a correspondence of each spectral basis element to a certain musical sound.

From the numerous potential applications, we concentrate on two to illustrate the proposed framework. We describe next the challenges in these applications. This early description of these applications will help to further motivate the work.

\subsubsection{Introduction to hyperspectral imaging (HSI)}
HSI sensors record up to several hundred different frequencies in the visible, near-infrared and infrared spectrum. This precise spectral information provides some insight on the material at each pixel in the image. Due to relatively low spatial resolution and the presence of multiple materials at a single location (e.g., tree canopies above ground or water and water sediments), many pixels in the image contain the mixed spectral signatures of multiple materials. The task of determining the {\it abundances} (presence quantity) of different materials in each pixel is called {\it spectral unmixing}. This is clearly an ill-posed problem that requires some assumptions and data models.

Unmixing requires a dictionary with the spectral signatures of the possible materials (often denoted as {\it endmembers}). Since these dictionaries can be difficult to obtain and might depend on the conditions under which they were recorded, it is sometimes desirable to automatically extract suitable endmembers from the image one wants to demix in a process called {\it endmember detection}. Many different techniques for endmember detection have been proposed, see \cite{Zare08} and references therein. Related although not yet applied to endmember detection are subset selection methods like the rank-revealing QR decomposition (e.g. \cite{Boutsidis09,CH}), which finds the most linearly independent columns of a matrix.\footnote{Independent of the work here described, Laura Balzano, Rob Nowak and Waheed Bajwa developed a related matrix factorization technique and connected and compared it to RRQR. We thank Laura for pointing out their work \cite{Balzano10} and also the possible relationships with RRQR.}  However, unlike our approach, QR methods do not take non-negativity constraints into account.  The general principle behind rank-revealing QR for subset selection is to find a column permutation of the data matrix such that the first few columns are as well conditioned as possible \cite{CH}.

Simultaneously detecting endmembers and computing abundances can be stated as factoring the data matrix $X \in \mathbb{R}^{m,d}$ into $X \approx AS$, $A,S \geq0$, with both $A \in \mathbb{R}^{m,n}$ and $S \in \mathbb{R}^{n,d}$ being unknown. In this notation each column of $X$ is the spectral signature of one pixel in the image. Hence, $m$ is the number of spectral bands, $d$ is the total number of pixels, and each of the $n$ columns of $A$ represents one endmember. The abundance matrix $S$ contains the amounts of each material in $A$ at each pixel in $X$.  The application of NMF to hyperspectral endmember detection can for instance be found in \cite{Miao07,BBLPP}.

Considering that while material mixtures in HSI exist, it is unlikely that pixels contain a mixture of all or many of the materials in $A$, researchers have recently focused on encouraging sparsity on the abundance matrix $S$ \cite{Arthurunmix, L1unmix}. Motivated by the ideas of dictionary learning for sparse coding, the works \cite{Greer10, Castrodad10} proposed explicitly to look for endmember matrices that lead to sparse abundance maps. We follow a similar idea, though our method will be fundamentally different in two aspects: First, we restrict columns of our dictionary $A$ to appear somewhere in the data $X$. This is a common working hypothesis for moderate ground sampling distance images and is called {\it pixel purity} assumption.  It corresponds to the partial orthogonality assumption on $S$ discussed previously.  In the general context of dictionary learning and non-negative matrix factorization it guarantees the columns of $A$ to be physically meaningful. As mentioned above, the lack of physical interpretation has been a critical shortcoming of standard dimensionality reduction and dictionary learning techniques, and not yet addressed in these areas of research.  Second, choosing the dictionary columns from the data enables us to propose a convex model and hence avoid the problem of saddle points or local minima.

Different areas of applications use different terminology for mathematically similar things. Throughout this paper we will use the terminology of hyperspectral unmixing to explain and motivate our model, although its application is not exclusively in the field of HSI. For instance, an endmember could be any abstract column of the dictionary matrix $A$ with a different physical meaning depending on its context. However, we think it aids the clarity and understanding of the model to use the HSI example throughout the explanations. To show that our model is not limited to the hyperspectral case we also present results for the problem of blind source separation, which we will briefly describe in the next subsection.

\subsubsection{Introduction to blind source separation (BSS)}
Blind source separation is the general problem of recovering unknown source signals from measurements of their mixtures, where the mixing process is unknown but often assumed to be linear.  Examples include demixing audio signals and nuclear magnetic resonance (NMR) spectroscopy.  Many BSS problems have the same linear mixing model $X = AS$ that we are using for hyperspectral endmember detection and unmixing, often also with a nonnegativity constraint on $S$ \cite{NN}.  Here the rows of $S$ represent the source signals, $A$ is the unknown mixing matrix and the rows of $X$ are the measured signals.  The analogy of the hyperspectral pixel purity assumption can also be applied to some BSS problems \cite{NN}.  The interpretation is that for each source there is some place where among all the sources only that source is nonzero.  Thus up to scaling, the columns of $A$ should appear somewhere in the columns of the data matrix $X$ and we can use the same algorithm we use for endmember detection.

\section{Convex endmember detection model \label{sec:method}}
\subsection{Convexification of matrix factorization using the pixel purity assumption}
As mentioned above, we are assuming at first that the endmembers can be found somewhere in the data $X$. This assumption will enable us to propose a convex model for factoring the data $X$ into a product $AS$, a problem usually tackled by non-convex optimization techniques.  We assume that there exists an index set $I$ such that the columns $X_i$ of $X$ are endmembers for $i \in I$. Under the assumption of non-negative linear mixing of signals, this means that any column $X_j$ in $X$ can be written as
\begin{eqnarray}
\label{puritycombi}
X_j = \sum_{i \in I} X_i T_{i,j},
\end{eqnarray}
for coefficients $T_{i,j} \geq 0$. The problem is that the coefficients $T_{i,j}$ as well as the index set $I$ are unknown. Hence, we start by using all columns in $X$ to describe $X$ itself, i.e. we look for coefficients $T\geq0$ for which
\begin{eqnarray}
\label{eq:XXT}
X=XT.
\end{eqnarray}
Notice that Equation (\ref{eq:XXT}) has many solutions. However, we know that the desired representation uses as few columns of $X$ as possible, i.e., only the endmembers. Since not using the $j^{th}$ column in the above formulation corresponds to having the entire $j^{th}$ row of $T$ be zero, we can reformulate the endmember detection problem as finding a solution to (\ref{eq:XXT}) such that as many rows of $T$ as possible are zero. Mathematically,
\begin{eqnarray}
\label{l0}
\min_{T\geq 0} \|T\|_{\text{row-}0} \text{  such that  } XT = X,
\end{eqnarray}
where $\|T\|_{\text{row-}0}$ denotes the number of non-zero rows. The columns of $X$ that correspond to non-zero rows of the minimizer $T$ of (\ref{l0}) are the desired endmembers (which define the lowest dimensional subspace where the data resides). Since problem (\ref{l0}) is not convex we relax the above formulation by replacing $\|T\|_{\text{row-}0}$ by the convex $l_{1,\infty}$ norm $\|T\|_{1,\infty} = \sum_i \max_j |T_{i,j}|$. The $l_1$ part of the norm should encourage sparsity. Combined with the maximum norm we can expect that the vector of maxima among each row becomes sparse which means we obtain row sparsity. Notice that we could have chosen other row-sparsity encouraging regularizers such as $l_{1,2}$. However, we will prove in Section \ref{sec:l0} the $l_{1,\infty}$ relaxation is exact in the case of normalized, non-repeating data, which makes it clearly preferable.

As mentioned in the previous section, it is important to take noise into account and therefore the equality (\ref{eq:XXT}) is too restrictive for real data. Hence, we will introduce a parameter that negotiates between having many zero rows (also called \textit{row sparsity}) and good approximation of the data $\|X-XT\|_F^2$ in the Frobenius norm.
Furthermore, for $T$ to be a good solution, not only should most of its rows be zero, but the nonzero rows should also be sparse, since sparsity of the coefficient matrix reflects physically reasonable prior knowledge. For hyperspectral unmixing the additional sparsity requirement on $T$ reflects the assumption that most pixels are not mixtures of all the selected endmembers, but rather just a few. Thus, we will add an additional weighted $l_1$ term to incorporate this second type of sparsity (see \cite{sprechmann10} for a model combining structured and collaborative sparsity with individual sparsity)
.

\subsection{Data reduction as preprocessing \label{sec:dimred}}
It is already clear from Equation (\ref{eq:XXT}) that the problem is too large because the unknown matrix $T$ is a $d \times d$ matrix, where $d$ is the number of pixels.  Thus, before proceeding with the proposed convex model, we reduce the size of the problem by using clustering to choose a subset of candidate endmembers $Y$ from the columns of $X$ and a submatrix $X_s \in \R^{m \times d_s}$ of $X$ for the data with $d_s \leq d$. In other words, we want to reformulate the problem as $YT \approx X_s$ with $X_s \in \R^{m \times d_s}$, $Y \in \R^{m \times n_c}$, $T \in \R^{n_c \times d_s}$ with $n_c<<d$, $d_s\leq d$. We use $X_s = Y$ in our experiments but could also include more or even all of the data.  We use k-means with a farthest-first initialization to select $Y$.  An angle constraint $\langle Y_i, Y_j \rangle < .995$ ensures the endmember candidates, namely the columns of $Y$, are sufficiently distinct.  An upper bound is placed on the number of allowable clusters so that the size of the problem is reasonable.  We then propose a convex model for the more manageable problem of finding a nonnegative $T$ such that $YT \approx X_s$, with $T$ having the same sparsity properties described above.  Note that we have not convexified the problem by pre-fixing the dictionary $Y$.  This is done simply to work with manageable dimensions and datasets.  Our convex model will still select the endmembers as a subset of this reduced dataset $Y$, namely the columns of $Y$ that will correspond to non-zero rows of $T$.

\subsection{The endmember selection model}
Our model consists of a data fidelity term and two terms that encourage the desired sparsity in $T$.  For simplicity, we consider the data fidelity term
\begin{eqnarray}
\frac{\beta}{2}\|(YT-X_s)C_w\|_F^2,
\end{eqnarray}
 where $\|\cdot\|_F$ denotes the Frobenius norm, $\beta$ is a positive constant, and $C_w \in \R^{d_s \times d_s}$ is a diagonal matrix we can use to weight the columns of $(YT-X_s)$ so that it reflects the density of the original data. As mentioned earlier we encourage rows of $T$ to be zero by penalizing with $\zeta \|T\|_{1,\infty}$, which for non-negative $T$ equals $\zeta \sum_i \max_j(T_{i,j})$, with $\zeta$ a positive constant, so that only a few samples are cooperatively selected as endmembers.  \footnote{The work mentioned above by Balzano, Nowak and Bajwa uses block orthogonal matching pursuit and also mentions the possible use of $\| \cdot \|_2$ instead of $\| \cdot \|_{\infty}$.}
 This kind of collaborative/structured sparsity regularizer has been proposed in several previous works, for example \cite{tropp06,MJOB}.

 To encourage sparsity of the nonnegative $T$, we use a weighted $l_1$ norm, $\langle R_w \sigma C_w , T \rangle$.  Here $R_w$ is a diagonal matrix of row weights.  We choose $R_w$ to be the identity in our experiments, but if for example we wanted to encourage selection of endmembers towards the outside of the cone containing the data, we could also choose these weights to be proportional to $\langle Y_j, \bar{Y} \rangle$, where $\bar{Y}$ is the average of the columns of $Y$. This would encourage the method to prefer endmebers further away from the average $\bar{Y}$. The weighting matrix $\sigma$ has the same dimension as $T$, and the weights are chosen to be
 \begin{eqnarray}
 \sigma_{i,j} = \nu (1 - e^{\frac{-(1-(Y^TX_s)_{i,j})^2}{2h^2}}),
 \end{eqnarray}
for constants $h$ and $\nu$.  This means that $\sigma_{i,j}$ is small when the $i$th column of $Y$ is similar to the $j$th column of $X_s$ and larger when they are dissimilar.  This choice of weights encourages sparsity of $T$ without impeding the effectiveness of the other regularizer.  Since the smallest weight in each row occurs at the most similar data, this helps ensure a large entry in each nonzero row of $T$, which makes the row sparsity term more meaningful.  It also still allows elements in a column of $T$ to be easily shifted between rows corresponding to more similar endmember candidates, which can result in a reduction of $\zeta \sum_i \max_j(T_{i,j})$ without significantly affecting the weighted $l_1$ term.  Since the weighted $l_1$ penalty here is really just a linear term, it can't be said to directly enforce sparsity, but by encouraging each column of $T$ to sum to something closer to one, the weighted $l_1$ penalty prefers data to be represented by nearby endmember candidates when possible, and this often results in a sparser matrix $T$.  Overall the proposed convex model is given by
\begin{equation}
\label{convex_model}
\min_{T \geq 0} \zeta \sum_i \max_j(T_{i,j}) + \langle R_w \sigma C_w , T \rangle + \frac{\beta}{2}\|(YT-X_s)C_w\|_F^2 .
\end{equation}
For our experiments we normalize the columns of $X$ to have unit $l_2$ norm so that we discriminate based solely on spectral signatures and not intensity.

\subsection{Refinement of solution}
Since we are using a convex model to detect endmembers, it cannot distinguish between identical or very similar endmember candidates, which we will discuss from a more theoretical point of view in Section \ref{sec:l0}.  However, the model works very reliably when the columns of $Y$ are sufficiently distinct, which they are by construction.  A limitation is that the convex model is unable to choose as endmembers any data not represented in $Y$.  Nonetheless, as is shown in Section \ref{sec:results}, the results of this approach already compare favorably to other methods.  Moreover, it provides an excellent initialization for the alternating minimization approach to NMF, which can be used to further refine the solution if desired.  Letting $\tilde{A}$ be the endmembers selected by the convex model, the solution is refined by alternately minimizing
\begin{equation}\label{alt} \min_{A \geq 0, S \geq 0, ||A_j - \tilde{A}_j\|_2 < a_j} \frac{1}{2}\|AS - X||_F^2 + \langle R_w \sigma , S \rangle \end{equation}
and renormalizing the columns of $A$ after each iteration. Here, $a_j$ is the diameter of the $j$th cluster containing the data near $\tilde{A}_j$, and ensures that the refined endmembers obtained by this alternating approach cannot be too different from those already selected by the convex model, thereby remaining as close as desired to the physical space.

To recover the full abundance matrix $S$ without refining $\tilde{A}$, we can solve the convex minimization problem in Equation (\ref{alt}) for $S$ using the full data matrix $X \in \R^{m \times d}$ and the original endmembers $\tilde{A} \in \R^{m \times n}$ selected by the convex model.

\subsection{Numerical optimization\label{sec:numerics}}
We use the alternating direction method of multipliers (ADMM) \cite{GM,GlM} to solve (\ref{convex_model}) by finding a saddle point of the augmented Lagrangian
\begin{align*}
L_{\delta}(Z,T,P)  =& g_{\geq 0}(T) + \zeta \sum_i \max_j(T_{i,j})\\
& + \langle R_w \sigma C_w , T \rangle  + \frac{\beta}{2}\|(YZ-X_s)C_w\|_F^2 \\
& + \langle P , Z - T \rangle + \frac{\delta}{2}\|Z-T\|_F^2 ,
\end{align*}
where $g_{\geq 0}$ is an indicator function for the $T \geq 0$ constraint.

In iteration $k+1$ the algorithm proceeds by minimizing first $L_{\delta}(Z,T^k,P^k)$ with respect to $Z$ to get $Z^{k+1}$, then minimizing $L_{\delta}(T,Z^{k+1},P^k)$ with respect to $T$ to get $T^{k+1}$, and then updating the Lagrange multiplier $P$ by $P^{k+1} = P^k + \delta (Z^{k+1} - T^{k+1})$.  Each minimization step is straightforward to compute, and the algorithm is guaranteed to converge for any  $\delta > 0$.

Note that it is faster to precompute and store the inverse involved in the update for $Z$, but this can be overly memory intensive if $Y$ has many columns and $C_w \neq I$.  One could use a more explicit variant of ADMM such as the method in \cite{ZBO} to avoid this difficulty if a larger number of columns of $Y$ is desired.  Here we have restricted this number $n_c$ to be less than $150$ and ADMM can be reasonably applied.

Also note that the minimization problem for $T^{k+1}$,
\begin{eqnarray}
 T^{k+1} &=& \arg \min_T g_{\geq 0}(T) + \zeta \sum_i \max_j(T_{i,j}) \nonumber \\
 && + \frac{\delta}{2}\|T - Z^{k+1} - \frac{P^k}{\delta} + \frac{R_w \sigma C_w}{\delta}\|_F^2 ,
 \end{eqnarray}
decouples into separate problems for each row $T_i$.  The Legendre transformation of $g_{\geq 0}(T_i) + \zeta \max_j(T_{i,j})$ is the indicator function for the set $C_{\zeta} = \{P_i \in \R^{d_s}: \|\max(P_i,0)\|_1 \leq \zeta \}$.
Let $\tilde{T}^{k+1} = Z^{k+1} + \frac{P^k}{\delta} - \frac{R_w \sigma C_w}{\delta}$.  Then by the Moreau decomposition \cite{Mo}, the minimizer $T^{k+1}$ is given by
\[T^{k+1} = \tilde{T}^{k+1} - \Pi_{C_{\frac{\zeta}{\delta}}}(\tilde{T}^{k+1}) , \]
where $\Pi_{C_{\frac{\zeta}{\delta}}}(\tilde{T}^{k+1})$ orthogonally projects each row of $\tilde{T}^{k+1}$ onto $C_{\frac{\zeta}{\delta}}$, which is something that can be computed with complexity $O(d_s \log(d_s))$.

The other algorithm parameters we use are $\delta = 1$, $\zeta = 1$, $\beta = 250$, $\nu = 50$, and $h = 1 - \cos(4 \pi /180)$.  In our experiments, we also choose $X_s = Y$.  We then define column weights $C_w$ that weight each column $j$ by the number of pixels in the $j$th cluster (the cluster centered at $Y_j$) divided by the total number of pixels $d$.  To refine the solution of the convex model, we note that each alternating step in the minimization of (\ref{alt}) is a convex minimization problem that can again be straightforwardly minimized using ADMM and its variants.  The update for the abundance $S$ is identical to the split Bregman algorithm proposed for hyperspectral demixing in \cite{Arthurunmix, L1unmix}, and its connection to ADMM is discussed in \cite{E}.

\section{The connection between row-$0$ and $l_{1,\infty}$ \label{sec:l0}}
In this section we will show that the motivation for our model comes from a close relation between the convex $l_{1,\infty}$ norm and the row-$0$ norm, which counts the number of non-zero rows.

\subsection{Distinct noise-free data}
Let us assume we have data $X$ which is completely noise free, normalized, obeys the linear mixing model, and contains a pure pixel for each material. Under slight abuse of notation let us call this data after removing points that occur more than once in the image, $X$ again. As discussed in Section \ref{sec:method} the endmember detection problem can now be reformulated as finding the minimizer $T$ of (\ref{l0}), where the true set of endmembers can then be recovered as the columns of $X$ that correspond to non-zero rows of $T$.

The fact that problem (\ref{l0}) gives a solution to the endmember selection problem is not surprising, since (\ref{l0}) is a non-convex problem and related problems (for instance in compressed sensing) are usually hard to solve. However, due to the non-negativity constraint we will show that $l_{1,\infty}$ minimization is an exact relaxation of the above problem.

For any $T\geq0$ with $XT=X$ the entries of $T$ are less than or equal to one, $T_{i,j}\leq1.$\footnote{This is a simple fact based on the normalization and non-negativity, $1 = \|X_k\| = \|\sum_i T_{i,k} X_i\| \stackrel{T,X \geq 0}{\geq} \max_i \| T_{i,k} X_i\| = \max_i T_{i,k}$} Furthermore, the endmembers can only be represented by themselves which means that in each row with index $i$, $i \in I$, we have a coefficient equal to 1. We can conclude that the $l_{1,\infty}$ norm of any $T \geq 0$ with $XT=X$ is
\begin{eqnarray}
\| T \|_{1,\infty} &=& \sum_{i=1}^{d} \max_j T_{i,j},\\
&\geq& \sum_{i \in I} \max_j T_{i,j} = |I|.
\end{eqnarray}
However, it is possible to have equality in the above estimate if and only if $T_{i,j}=0$ for $i \notin I$. In this case the rows of the non-negative $l_{1,\infty}$ minimizer of $XT=X$ are only supported on $I$, which means it is a minimizer to the row-$0$ problem (\ref{l0}). Vice versa, any row-$0$ minimizer $\hat{T}$ has exactly one entry equal to one in any row corresponding to an endmember and zero rows elsewhere, which means $\|\hat{T}\|_{1,\infty}=|I|$, and hence $\hat{T}$ also minimizes the $l_{1,\infty}$ norm under the $XT=X$ constraint. We therefore have shown the following lemma:
\begin{lemma}
If we remove repeated columns of $X$ and have normalized data, the sets of minimizers of
\begin{eqnarray}
\min_{T\geq0} \| T \|_{\text{row-}0} \text{  such that  } XT=X
\end{eqnarray}
and
\begin{eqnarray}
\min_{T\geq0} \| T \|_{1,\infty} \text{  such that  } XT=X
\end{eqnarray}
are the same. 
\end{lemma}
Notice that while generally there are other regularizations, like for instance $l_{1,2}$, which would also encourage row sparsity, this lemma only holds for $l_{1,\infty}$. For $XT=X$, the $l_{1,\infty}$ norm counts the number of rows and is not influenced by any value that corresponds to a mixed pixel. This property is unique for $l_{1,\infty}$ regularization with a non-negativity constraint and therefore makes it the preferable choice in the proposed framework.

\subsection{Noise in the data}
Of course the assumptions above are much too restrictive for real data. Therefore, let us look at the case of noisy data of the form $X^\delta = X + N$, where $X$ is the true, noise-free data with no repetition of endmembers as in the previous section, and $N$ is noise bounded in the Frobenius norm by $\|N\|_F \leq \delta$. We now consider the model
\begin{eqnarray}
\label{eq:simpleenergy}
J_\alpha(T) = \|XT - X\|_F^2 + \alpha \|T\|_{1,\infty} \text{  such that } T\geq 0.
\end{eqnarray}
The following lemma shows that for the right, noise-dependent choice of regularization parameter $\alpha$, we converge to the correct solution as the noise decreases.
\begin{lemma}
\label{noisezeroconv}
Let $\hat{T}$ be a non-negative $\| \cdot \|_{1,\infty}$-minimum norm solution of $XT=X$, $X^\delta = X + N$ be noisy data with $\|N\|_F \leq \delta$ and $T_\alpha^\delta$ denote a minimizer of energy functional (\ref{eq:simpleenergy}) with regularization parameter $\alpha$ and replacing $X$ by $X^\delta$. If $\alpha$ is chosen such that
\begin{eqnarray}
\alpha \rightarrow 0, \ \ \frac{\delta^2}{\alpha}\rightarrow 0 \ \ \ \text{as } \delta \rightarrow 0
\end{eqnarray}
then there exists a convergent subsequence $T_{\alpha_n}^{\delta_n}$ and the limit of each convergent subsequence is a $\| \cdot \|_{1,\infty}$-minimum norm solution of $XT=X$.
\end{lemma}
\noindent For the sake of clarity we moved the proof of this lemma to the appendix.
Lemma \ref{noisezeroconv} shows that our regularization is stable, since for decreasing noise and appropriate choice of the parameter $\alpha$ we converge to a non-negative $\| \cdot \|_{1,\infty}$-minimum norm solution of $XT=X$ which - as we know from the first part - gives the true solution to the endmember detection problem \emph{when the columns of $X$ are distinct}.
While identical points are easy to identify and eliminate in the noise free data, determining which points belong to the same endmember in the noisy data can be very difficult. This is the reason for our first data reduction step. Lemma \ref{noisezeroconv} tells us that for our method to be closely related to the row-$0$ approach we have to avoid having several noisy versions of the same endmember in $X$. We therefore impose an angle constraint as described in Section \ref{sec:dimred} while clustering the data, which basically corresponds to an upper bound on the noise and states up to which angle signals might correspond to the same point.

\section{Numerical results \label{sec:results} for HSI}
In this section we present numerical results on endmember detection for supervised and real hyperspectral data and compare our results to existing detection algorithms. Since the goal of the paper is to present a general new convex framework for matrix factorization which is applicable to multiple areas, the hyperspectral numerical results are intended mainly as a proof of concept.  It is therefore encouraging that our method  is competitive with some of the established methods it is compared to in the examples below.

\subsection{Application to blind hyperspectral unmxing}
\subsubsection{Supervised endmember detection\label{sec:supenddet}}
\begin{figure*}[htb]
  \centering
  \centerline{\includegraphics[width=17cm]{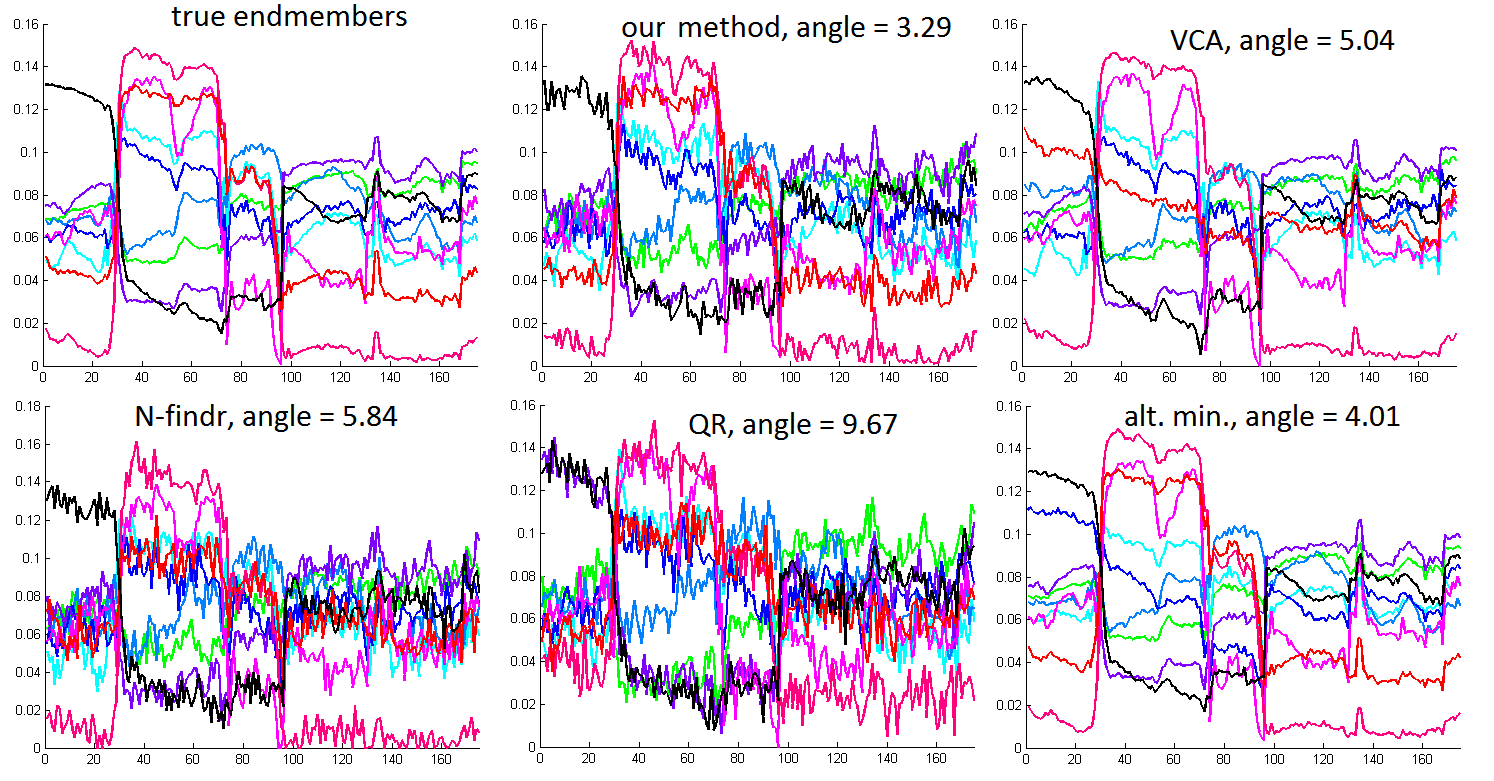}}
  \caption{Comparison of endmember recostruction methods}
  \label{indiansignals}
\end{figure*}

For comparison purposes we extracted nine endmembers from the standard Indian pines dataset (publicly available at \url{https://engineering.purdue.edu/ biehl/MultiSpec/hyperspectral.html}) by averaging over the corresponding signals in the ground truth region. Then we created 50 data points for each endmember, 30 data points for each combination of two different endmembers, 10 data points for each combination of three different endmembers, and additionally 30 data points as mixtures of all endmembers. Finally, we add Gaussian noise with zero mean and standard deviation 0.006, make sure our data is positive, and normalize it. We evaluate our method in a comparison to N-findr \cite{winter99}, vertex component analysis (VCA) \cite{Nascimento07, Nascimento05} with code from \cite{toolbox}, an NMF method using the alternating minimization scheme of our refinement step with random initial conditions, and the QR algorithm. For the latter we simply used MATLAB's QR algorithm to calculate a permutation matrix $\Pi$ such that $X \Pi = QR$ with decreasing diagonal values in $R$ and chose the first nine columns of $X \Pi$ as endmembers. Since the success of non-convex methods depends on the initial conditions or on random projections, we run 15 tests with the same general settings and record the average, maximum and minimum angle by which the reconstructed endmember vectors deviate from the true ones, see Table \ref{comparison}. For the tests we adjusted the parameters of our method to obtain 9 endmembers.
\begin{minipage}{0.5 \textwidth}
\vspace{0.2cm}
\captionsetup{type=table}
	\centering
		\begin{tabular}{|c|c|c|c|}
		\hline
			\textbf{Method} & \multicolumn{3}{|c|}{\textbf{Evaluation on 15 test runs}} \\
			\hline
			  & Avg. $\alpha$ & Min. $\alpha$ & Max. $\alpha$ \\
			\hline
			\hline
			Ours refined &  \textbf{3.37} & 3.30 & \textbf{3.42} \\
			\hline
			Ours without refinement & 3.93  & 3.84 & 4.01 \\
			\hline
			VCA  & 4.76 & 1.78 & 6.95 \\
			\hline
			N-findr & 10.19 & 7.12  &13.79 \\
			\hline
			QR & 9.87 & 4.71  & 12.74  \\
			\hline
			Alt. Min. & 4.50 & \textbf{1.76} & 8.17 \\
			\hline
		\end{tabular}
	\caption{Deviation angle from true endmembers}
	\label{comparison}
\vspace{0.2cm}	
\end{minipage}
We can see that our method gives the best average performance. Due to a high noise level, methods that rely on finding the cone with maximal volume or finding most linearly independent vectors, will select outliers as the endmembers and do not yield robust results. Looking at the minimal and maximal $\alpha$ we see the effect predicted. The non-convex methods like alternating minimization and VCA can outperform our method on some examples giving angles as low as 1.76. However, due to the non-convexity they can sometimes find results which are far off the true solution with deviation angles of 6.95 or even 8.17 degrees. A big benefit of our convex framework is that we consistently produce good results. The difference between the best and the worst reconstruction angle deviation for our method is 0.15 degrees with and 0.17 without the refinement, which underlines its high stability. Figure \ref{indiansignals} shows the original endmembers as well as an example reconstructions by each method with the corresponding angle of deviation.

\subsubsection{Results on real hyperspectral data}
\begin{figure*}[htb]
  \centering
  \centerline{\includegraphics[width=17cm]{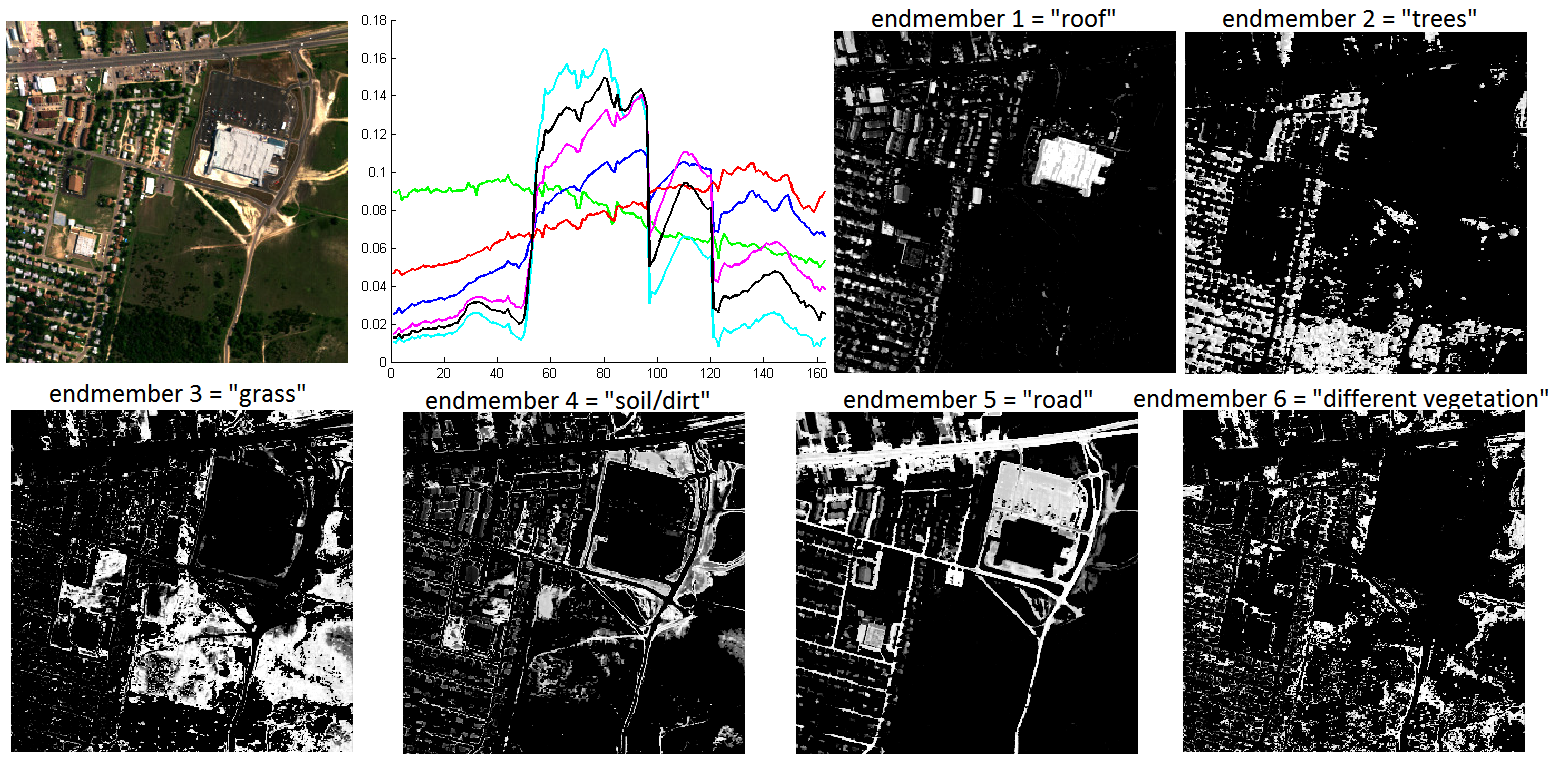}}
  \caption{Results on the urban hyperpectral image. First row, left: RGB image to compare the fraction planes to. First row, middle left: spectral signatures of the endmembers our method found. First row middle right, right and second row: abundance maps of the six endmembers.}
  \label{urbanresult}
\end{figure*}

To show how our method performs on real hyperspectral data we use the urban image (publicly available at \url{www.tec.army.mil/hypercube}). Figure \ref{urbanresult} shows the RGB image of the urban scene, the spectral signatures of the endmembers our method extracted, and the abundance maps of each endmember, i.e., each row of $T$ written back into an image.

The abundance maps are used in HSI to analyze the distribution of materials in the scene. First of all our method managed to select six endmembers from the image, which is a very reasonable dimension reduction for hyperspectral image analysis. We can see that the abundance maps basically segment the image into clusters of different material categories such as concrete, house roofs, soil or dirt, grass, and two different types of vegetation, which all seem to be reasonable when visually comparing our results to the RGB image. The spectral signatures our method selected are relatively distinct, but do not look noisy. Furthermore, the abundance maps seem to be sparse which was a goal of our method and reflects the physically reasonable assumption that only very few materials should be present at each pixel. 

\begin{figure*}[htb]
  \centering
  \centerline{\includegraphics[width=17cm]{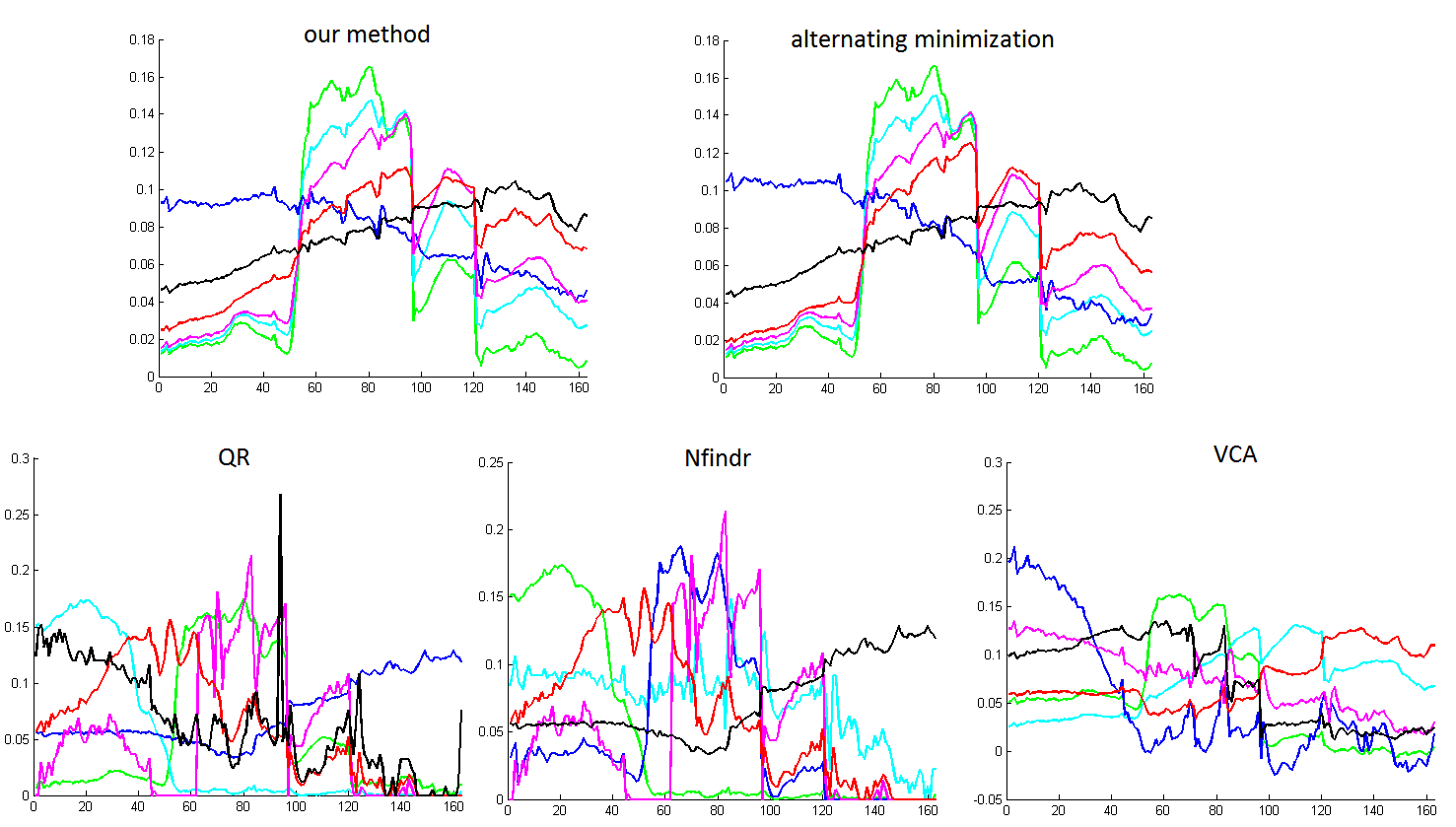}}
   \caption{Spectral signatures of endmembers extracted by different methods. Top row: results of our method and the alternating minimization approach. Bottom row: endmembers found by N-findr, QR and VCA}
  \label{fig:differentendmembersignals}
\end{figure*}

As a next step we compare our results to the ones obtained by N-findr, VCA, QR and alternating minimization. Unfortunately, we have no ground truth for the urban image, which is why we will look at the non-negative least squares (NNLS) unmixing results based on the endmembers $A_m$ each method found. Notice that geometrically NNLS gives the projection of the data onto the cone spanned by the endmembers. If $S_m$ denotes the NNLS solution of each method, the error $\|A_mS_m-X\|_F^2$ gives some insight on how much data is contained in the cone and therefore, how well the endmembers describe the data. However, as discussed earlier, due to noise we might not be interested in detecting the pixels furthest outside to be endmembers, although they might describe the data better. Thus, we will also report the sparsity of each cone projection $S_m$. Since any outside point will be projected onto a face or an edge of the cone, the sparsity will give some insight into whether the endmember vectors are well located. The more the endmembers are towards the inside of a point cluster, the more points we expect to be projected onto an edge of the cone rather than a face. Thus, a high sparsity relates to a reasonable position of an endmember. Table \ref{table:urbancomp} shows the relative number of non-zero coefficients, i.e., $\|S_m\|_0$ divided by the the total number of entries in $S_m$, as well as the projection error for N-findr, QR, VCA, alternating minimization and our method. Figure \ref{fig:differentendmembersignals} shows the corresponding endmember signals each method found.

\begin{table}[htbp]
\begin{tabular}{|l|r|r|r|r|r|}
\hline
 & \multicolumn{1}{l|}{Ours} & \multicolumn{1}{l|}{N-findr} & \multicolumn{1}{l|}{VCA} & \multicolumn{1}{l|}{QR} & \multicolumn{1}{l|}{Alt. Min.} \\ \hline
$\|A_mS_m - X\|^2$ & 533.9 & 4185.8 & 2516.5 & 1857.6 & \textbf{454.3} \\ \hline
$\|S_m\|_0/(d \cdot n)$& \textbf{0.40} & 0.60 & 0.47 & 0.60 & 0.41 \\ \hline
\end{tabular}
\caption{Comparison of different endmember detection methods in terms of error and sparsity of the data projection onto the cone spanned by the endmembers on the urban image. The corresponding endmember signatures are shown in Figure \ref{fig:differentendmembersignals}.}
\label{table:urbancomp}
\end{table}

We can see from the projection error that the sparse alternating minimization approach and our method found much better representation for the data than N-findr, VCA and QR, with the alternating minimization performing slightly better than our approach. Furthermore, the sparsity of the projection is also higher which indicates more reasonable choices for the endmembers. Looking at the spectral signatures in Figure \ref{fig:differentendmembersignals} we can speculate why this is the case. QR as well as N-findr chose very distinct signatures, which are probably far outliers in the dataset. Some of the VCA endmembers look even more extreme and take negative values, which is clearly unphysical and does not allow these endmembers to be interpreted as any material. On the other hand the endmembers of the alternating minimization approach and of our method are very similar and look comparably smooth. The average angle of deviation between our method and the alternating minimization approach is only 3.4 degrees, which lets us conclude that they basically converged to the same answer, which is encouraging considering the fact that these endmembers describe the rest of the data more than three times more accurately than the endmembers found by other methods. Furthermore, any signal our method selected differs at most by 0.036 degrees from an actual point in the data and is therefore physically meaningful.

\section{An extended model \label{sec:extended}}
We also propose an extended version of the model that takes into account the normalization of the data in order to better distinguish between noise and outliers.  We show this slightly more complicated functional can still be efficiently solved using convex optimization.

\subsection{Error model}
\begin{figure}[H]
\begin{center}
\epsfig{file=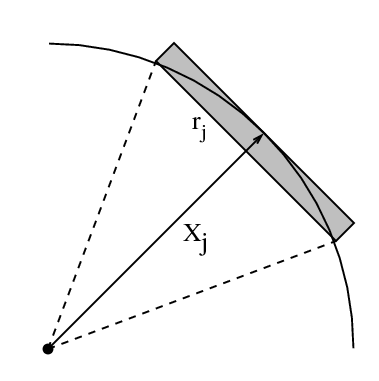,width=5cm,clip=}
\caption{Region of possible values for $X_j + V_j$ \label{hockey_puck}}
\end{center}
\end{figure}

We continue to work with the reduced set of endmember candidates $Y$ and a subset of the data $X_s$, which in practice we take to be $Y$.  Instead of penalizing $\|(YT-X_s)C_w\|_F^2$, we impose the linear constraint $YT-X_s = V - X_s \diag(e)$, where $T$,$V$ and $e$ are the unknowns.  $T$ has the same interpretation as before, $V \in \R^{m \times d_s}$ models the noise, and $e \in \R^{d_s}$ models the sparse outliers.  Since the columns of $X_s$ are normalized to have unit $l_2$ norm, we would also like most of the columns of $YT$ to be approximately normalized.  In modeling the noise $V$, we therefore restrict it from having large components in the direction of $X_s$.  To approximate this with a convex constraint $V \in D$, we restrict each column $V_j$ to lie in a hockey puck shaped disk $D_j$. Decompose $V_j = V_j^{\perp} + V_j^{\parallel}$, where $V_j^{\parallel}$ is the orthogonal projection of $V_j$ onto the line spanned by $X_j$ and $V_j^{\perp}$ is the radial component of $V_j$ perpendicular to $V_j^{\parallel}$.  Then given $0 \leq r_j < 1$, we restrict $\|V_j^{\perp}\|_2 \leq r_j$ and $\sqrt{1-r_j^2}-1 \leq V_j^{\parallel} \leq 0$.  The orthogonal projection onto this set is straightforward to compute since it is a box constraint in cylindrical coordinates.  This constraint set for $V_j$ is shown in Figure \ref{hockey_puck} in the case when $e_j = 0$.

We also allow for a few columns of the data to be outliers.  These are columns of $X$ that we don't expect to be well represented as a small error plus a sparse nonnegative linear combination of other data, but that we also don't want to consider as endmembers.  Given some $\gamma \geq 0$, this sparse error is modeled as $-X_s \diag{e}$ with $e$ restricted to the convex set $E = \{ e : e \geq 0 \text{ and } \sum_j (C_w e)_j \leq \gamma \}$.  Since $E$ is the nonnegative region of a weighted $l_1$ ball, the orthogonal projection onto $E$ can be computed with $O(d_s \log(d_s))$ complexity.  Here, since the weights $w_j$ sum to one by definition, $\gamma$ can be roughly interpreted as the fraction of data we expect to be outliers.  For non-outlier data $X_j$, we want $e_j \approx 0$ and for outlier data we want $e_j \approx 1$.  In the latter outlier case, regularization on the matrix $T$ should encourage the corresponding column $T_j$ to be close to zero, so $\|YT_j\|_2$ is encouraged to be small rather than close to one.

Keeping the $l_{1,\infty}$ regularization, the nonnegativity constraint and the weighted $l_1$ penalty from Equation (\ref{convex_model}), the overall extended model is given by

\begin{align}\label{convex_model2}\min_{T \geq 0, V_j \in D_j, e \in E} & \zeta \sum_i \max_j(T_{i,j}) + \langle R_w \sigma C_w , T \rangle \\
& \text{such that} \qquad YT-X_s = V - X_s \diag(e) . \notag \end{align}
The structure of this model is similar to the robust PCA model proposed in \cite{CLMW} even though it has a different noise model and uses $l_{1,\infty}$ regularization instead of the nuclear norm.

\subsection{Numerical optimization}
Since the convex functional for the extended model (\ref{convex_model2}) is slightly more complicated, it is convenient to use a variant of ADMM that allows the functional to be split into more than two parts.  The method proposed by He, Tao and Yuan in \cite{HTY} is appropriate for this application.  Again introduce a new variable $Z$ and the constraint $Z=T$.  Also let $P_1$ and $P_2$ be Lagrange multipliers for the constraints $Z-T=0$ and $YZ - V - X_s + X_s \diag(e) = 0$ respectively.  Then the augmented Lagrangian is given by
\begin{align*}
L_{\delta}(Z,T,V,e,P_1,P_2)  =&  g_{\geq 0}(T) + g_D(V) + g_E(e)  \\
& +\zeta \sum_i \max_j(T_{i,j}) + \langle R_w \sigma C_w , T \rangle  \\
& +\langle P_1 , Z - T \rangle \\
& + \langle P_2 , YZ - V - X_s + X_s \diag(e) \rangle  \\
&+ \frac{\delta}{2}\|Z-T\|_F^2\\
& + \frac{\delta}{2}\|YZ - V - X_s + X_s \diag(e)\|_F^2,
\end{align*}
where $g_D$ and $g_E$ are indicator functions for the $V \in D$ and $e \in E$ constraints.

Using the ADMM-like method in \cite{HTY}, a saddle point of the augmented Lagrangian can be found by iteratively solving the following subproblems with parameters $\delta > 0$ and $\mu > 2$,

\begin{align*}
Z^{k+1} & = \arg \min_Z  \left\| \bbm I \\ Y \ebm Z - \bbm T^k \\ V^k - X_s\diag(e^k) + X_s \ebm \right.\\
& \left. + \frac{1}{\delta} \bbm P_1^k \\ P_2^k \ebm \right\|_F^2 \\
T^{k+1} & = \arg \min_T g_{\geq 0}(T) + \zeta \sum_i \max_j(T_{i,j}) \\
& + \frac{\delta \mu}{2}\left\|T - T^k - \frac{1}{\mu}(Z^k - T^k) - \frac{P_1}{\delta \mu} + \frac{R_w \sigma C_w}{\delta \mu}\right\|_F^2\\
V^{k+1} & = \arg \min_V g_D(V) + \frac{\delta \mu}{2}\left\|V - V^k \right. \\
& \left. - \frac{1}{\mu}(YZ^{k+1} - V^k + X_s\diag(e^k) - X_s) - \frac{P_2^k}{\delta \mu} \right\|_F^2
\end{align*}
\begin{align*}
e^{k+1} & = \arg \min_e g_E(e) + \frac{\delta \mu}{2} \sum_{j=1}^{d_s} \Big( e_j - e_j^k + \frac{1}{\delta \mu} \sum_{i=1}^{m} \\
& (X_s)_{i,j}(P_2^k + \delta(YZ^{k+1} - V^k + X_s\diag(e^k) - X_s))_{i,j} \Big)^2 \\
P_1^{k+1} & = P_1^k + \delta (Z^{k+1} - T^{k+1}) \\
P_2^{k+1} & = P_2^k + \delta (YZ^{k+1} - V^{k+1} - X_s + X_s \diag(e^{k+1}))
\end{align*}

Each of these subproblems can be efficiently solved.  There are closed formulas for the $Z^{k+1}$ and $V^{k+1}$ updates, and the $e^{k+1}$ and $T^{k+1}$ updates both involve orthogonal projections that can be efficiently computed.

\subsection{Effect of extended model \label{sec:rgbexp}}
A helpful example for visualizing the effect of the extended model (\ref{convex_model2}) is to apply it to an RGB image.  Even though the low dimensionality makes this significantly different from hyperspectral data, it's possible to view a scatter plot of the colors and how modifying the model parameters affects the selection of endmembers.  The NMR data in Section \ref{sub:NMR} is four dimensional, so low dimensional data is not inherently unreasonable.

\begin{figure*}[htb]
\begin{center}

$\begin{array}{lll}
 \multicolumn{1}{c}{\text{original image}} &
\multicolumn{1}{c}{\nu=0, \gamma=0}&
\multicolumn{1}{c}{\nu=50, \gamma=.005} \\
\includegraphics[width=5.3cm]{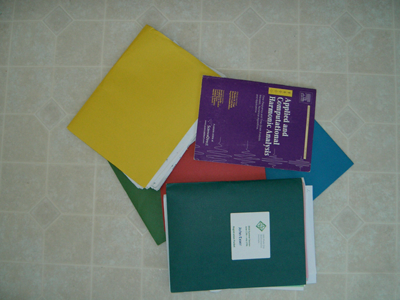} &
\includegraphics[width=5.3cm]{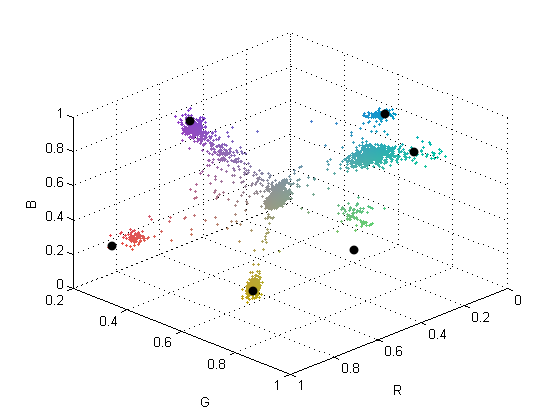} &
\includegraphics[width=5.3cm]{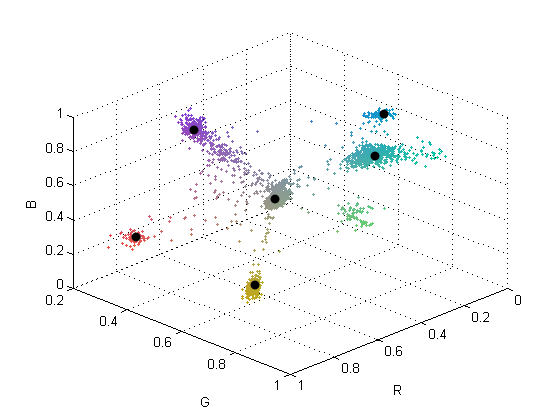}\\
\ &  & \\
\multicolumn{1}{c}{\text{T for } \nu=50, \gamma=.005} &
 \multicolumn{1}{c}{\nu=50, \gamma=.01} &
  \multicolumn{1}{c}{\nu=50, \gamma=.1}\\
\includegraphics[width=5.3cm]{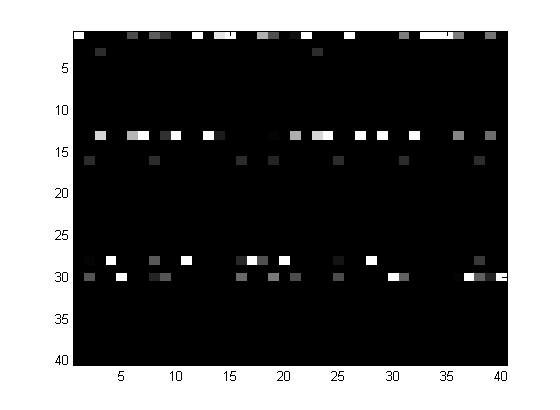} &
\includegraphics[width=5.3cm]{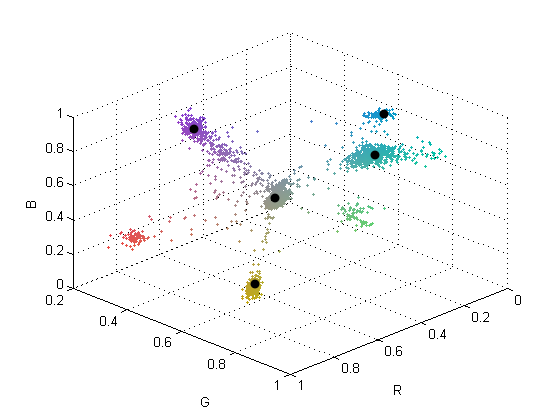} &
\includegraphics[width=5.3cm]{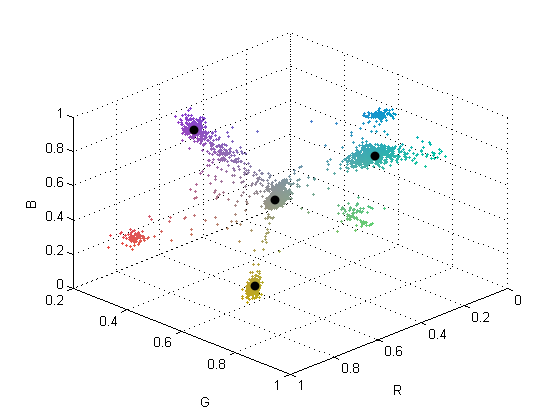}  \\
\ &  & \\
\end{array}$

\end{center}
\caption{Results of the extended model applied to RGB image. Upper left: RGB image we apply the blind unmixing algorithm to. Upper middle: 3d plot of the data points in the image in their corresponding color. Shown as black dots are the endmembers detected without allowing outliers ($\gamma=0$) and without encouraging particular sparsity on the coefficients ($\nu=0$). Upper right: With allowing some outliers the method removed an endmember in the one of the outside clusters, but included the middle cluster due to the encouraged sparsity. Lower left: Endmember coefficients for the parameter choice $\nu=50, \gamma=0.005$, where the brightness corresponds to the coefficient value. We can see that the coefficient matrix is sparse. Lower middle: Increasing the allowed outliers the red cluster endmember is removed. Increasing the outliers even further leads to decreasing the number of endmembers to 4. \label{folders}}
\end{figure*}

For the following RGB experiments, we use the same parameters as described in Section \ref{sec:numerics} and use the same k-means with farthest first initialization strategy to reduce the size of the initial matrix $Y$.  We do not however perform the alternating minimization refinement step.  Due to the different algorithm used to solve the extended model, there is an additional numerical parameter $\mu$, which for this application must be greater than two according to \cite{HTY}.  We set $\mu$ equal to $2.01$.  There are also model parameters $r_j$ and $\gamma$ for modeling the noise and outliers.  To model the small scale noise $V$, we set $r_j = \eta + m_j$, where $\eta$ is fixed at $.07$ and $m_j$ is the maximum distance from data in cluster $j$ to the cluster center $Y_j$.  To model the sparse error $e$, we use several different values of $\gamma$, which should roughly correspond to the fraction of data we are allowed to ignore as outliers.  We will also use different values of $\nu$, setting $\nu = 50$ to encourage a sparser abundance matrix and setting $\nu=0$ to remove the weighted $l_1$ penalty from the model.  Figure \ref{folders} shows the data, which is a color image of multicolored folders, the selected endmembers (black dots) from four different experiments, and the sparse abundance matrix $T$ for one of the experiments.

Note that in the $\nu=0$ experiment, sparsity of $T$ is not encouraged, so the dense gray cluster in the center of the cone, which corresponds to the floor in the image, is not selected as an endmember.  Additionally, the selected endmembers are towards the very outside of the cone of data.  In all the other experiments where $\nu=50$ the floor is selected as an endmember even though it is in the center of the cone of data.  This results in a much sparser matrix $T$.  Moreover, the selected endmembers tend to be in the center of the clusters of colors that we see in the scatter plot of the folder data.  Finally we note that as we increase the parameter $\gamma$, fewer endmembers are selected and some of the smaller outlying color clusters are ignored.

\subsection{Comparison between the base and the outlier model}
\begin{figure*}[htb]
  \centering
  \centerline{\includegraphics[width=15cm]{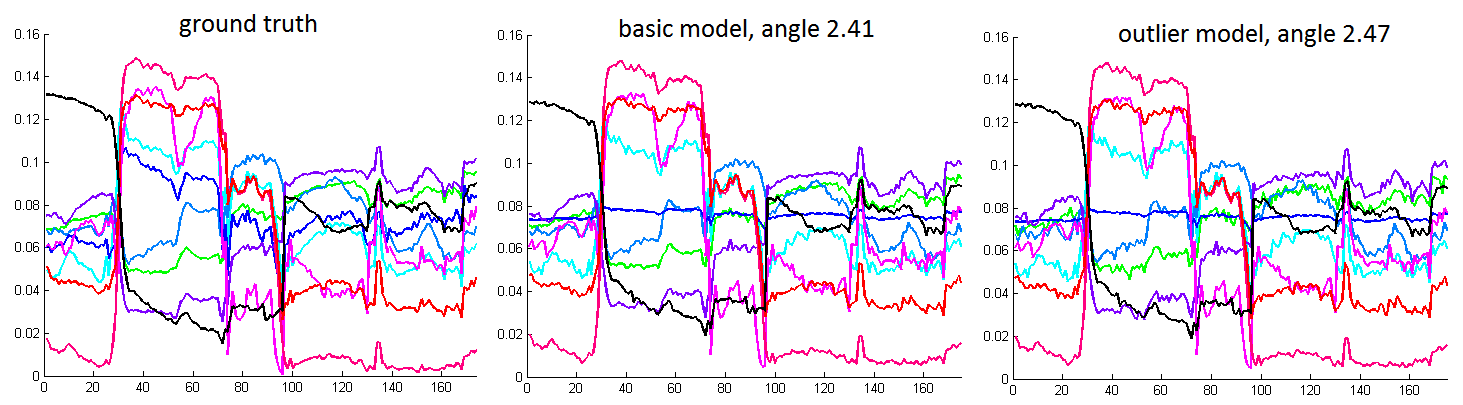}}
  \caption{Comparison between the basic and the outlier method on Indian pines data with Gaussian noise}
  \label{fig:comp1}
\end{figure*}

To illustrate the difference between the outlier and the base model we use the same dataset as for the supervised endmember detection experiment and first repeat the experiment from Section \ref{sec:supenddet} with 30 data points for each endmember, 20 data points for each combination of two different endmembers, 10 data points for each combination of three different endmembers, and additionally 30 data points as mixtures of all endmembers. We add Gaussian noise with zero mean and standard deviation 0.005, run the outlier model with $\zeta=1$, $\eta=0.08$, $\gamma=0.01$, $\nu=40$ and run the basic model with $\zeta=1.3$, $\nu=40$, $\beta=250$. Figure \ref{fig:comp1} shows the results for both methods with their average angle of deviation from the true endmembers. 

We can see that both models give good results close to the ground truth. Due to the Gaussian noise we added, the best possible choice of columns of $X$ deviates by an average angle of 3.29 degrees from the true endmembers. Due to the refinement step both methods could achieve an average angle below this value. The main remaining deviation is mainly due to the methods selecting an almost straight line rather then the corresponding true endmember.

\begin{figure*}[htb]
  \centering
  \centerline{\includegraphics[width=14cm]{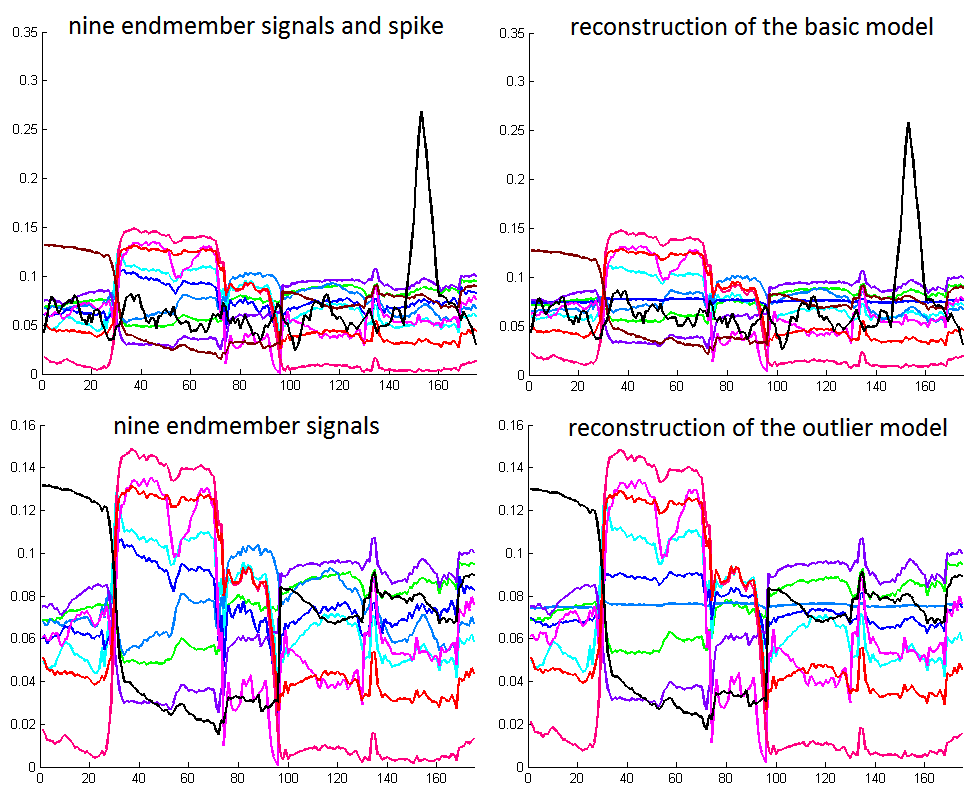}}
  \caption{Comparison between the basic and the outlier method on Indian pines data with outliers}
  \label{fig:comp2}
\end{figure*}

As a second experiment we want to simulate outliers in the data. First we again create a mixed pixel data set with nine endmembers as above, but without adding Gaussian noise. Next, we create a spike signal and add the spike itself as well as about $3\%$ data that is mixed with the spike signal. This shall simulate a small fraction of the data being outliers. Again, we run the basic as well as the outlier model on this dataset and obtain the results shown in Figure \ref{fig:comp2}. The upper left image shows the true nine endmembers plus the spike signal we used to create the outliers with. As we can see in the image on the upper right, which shows the result of the basic model, the algorithm selected the spike as an endmember. This is very reasonable because although only $3\%$ of the data contains parts of the spike, it is a strong outlier and hence expensive for the fidelity term in the Frobenius norm to exclude. The shown ten detected endmembers deviate only by 3.6 degrees from the nine true endmembers and the spike.

The second row of Figure \ref{fig:comp2} shows the nine true endmembers on the left and the result of the outlier model on the right. As we can see the outlier model was able to handle the small fraction of signals containing the spike and only selected the nine true endmembers. The average angle of deviation in this case is 2.7 degrees and we can confirm that the model behaves like we expected it to.

\subsection{Application to blind source separation of NMR data \label{sub:NMR}}
\begin{figure*}[htb]
  \centering
  \centerline{\includegraphics[width=12cm]{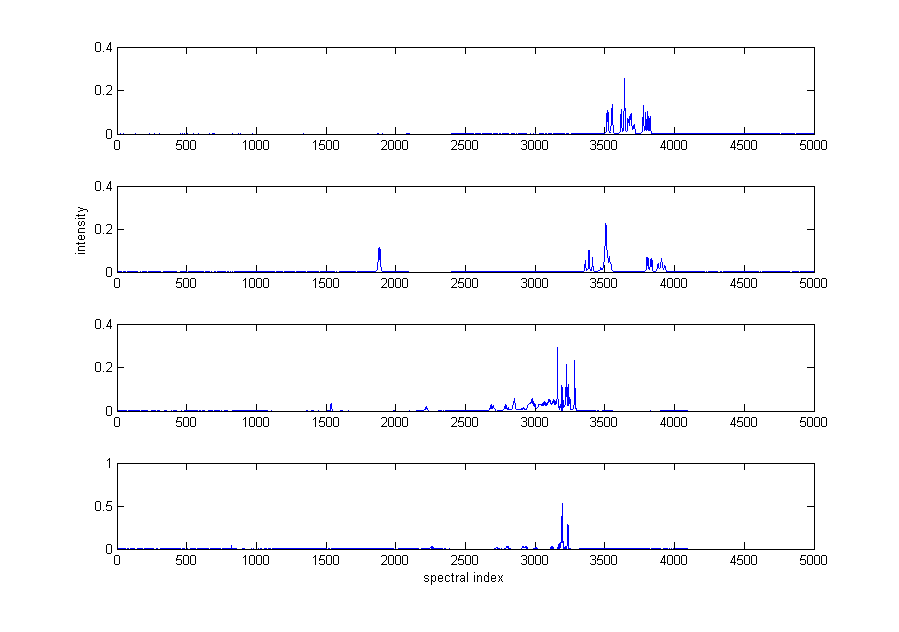}}
  \caption{Four NMR source spectra from \cite{NN} Fig. 4}
  \label{fig:sources}
\end{figure*}

To illustrate how our model can be applied to BSS problems, we use it to recover the four NMR source spectra from (\cite{NN} Fig. 4) from four noise free mixtures.  The four sources are shown in Figure \ref{fig:sources}.
Let $S_0 \in \R^{4 \times 5000}$ be the sources and let the mixtures $X_0$ be generated by $X_0 = A_0 S_0$ with
\[ A_0 = \bbm .3162 & .6576 & .3288 & .5000 \\
.3162 & .3288 & .6576 & .5000 \\
.6325 & .1644 & .1644 & .5000 \\
.6325 & .6576 & .6576 & .5000 \ebm . \]

We will use the outlier model to recover the mixing matrix $A$ from $X_0$.  Unlike the hyperspectral examples, some columns of $X_0$ here can be nearly zero if all sources are simultaneously zero at the same spectral index.  We can see from Figure \ref{fig:sources} that this is indeed the case.  Since our algorithm uses normalized data, we first remove columns of $X_0$ whose norm is below some threshold, which we take to be $.01 \max_j \|X_j\|$.  We then normalize the remaining columns to get $X$.  This simple approach suffices for this example, but in general the parameters $r_j$ for the $V \in D$ constraint could also be modified to account for columns of $X_0$ that have significantly different norms.

A minor difficulty in applying our method to this BSS problem is that we know $A$ should have four columns but there is no way to constrain the algorithm to produce a dictionary with exactly four elements.  We therefore adjust parameters until the dimension of the result is correct.  This is straightforward to do and could be automated.  For example, to choose a smaller dictionary, we can reduce $\nu$ and/or increase $\gamma$.

The parameters used here are identical to those used in the RGB experiments of Section \ref{sec:rgbexp} except $\gamma=.01$ and $\nu = 5$.  Also, for the data reduction step, the angle constraint was increased to $\langle Y_i , Y_j \rangle < .998$.  The computed mixing matrix after permutation is
\[ A = \bbm  .3267  &  .6524  &  .3327  &  .4933 \\
    .3180  &  .3300 &   .6544 &   .5110 \\
    .6228  &  .1757 &   .1658  &  .4836 \\
    .6358  &  .6593  &  .6585  &  .5114 \ebm . \]
Note that the columns of $A$ are normalized because the algorithm selects dictionary elements from a normalized version of the data $X$.  Since for this simple problem, $A$ is invertible, it is straightforward to recover the sources by $S = \max(0,A^{-1}X_0)$.  More generally, we can recover $S$ by minimizing the convex functional in Equation (\ref{alt}) with respect to $S$ using the un-normalized data matrix $X_0$ and the computed endmembers $A$.

\section{Future research}
\label{sec:future}
We have presented a convex method for factoring a data matrix $X$ into a product $AS$ with $S\geq0$ under the constraint that the columns of $A$ appear somewhere in the data $X$. This type of factorization ensures the physical meaning of the dictionary $A$, and we have successfully applied it to hyperspectral endmember detection and blind source separation in NMR. For non-repeating noise free data, the $l_{1,\infty}$ regularization was proven to be an exact relaxation of the row-$0$ norm. We further proposed an extended model that can better handle outliers. Possible future application areas include computational biology, sensor networks, and in general dimensionality reduction and compact representation applications where the physical interpretation of the reduced space is critical.  It will also be interesting to try and extend our convex model to the class of problems discussed in \cite{SRRSX} for which the pixel purity or non-overlapping assumption is approximately but not exactly satisfied.

\vspace{0.2cm}
\noindent
{\bf Acknowledgments - }
We would like to acknowledge Laura Balzano and Rob Nowak, who are working on a similar selection framework without non-negativity constraints, partially motivated by problems in sensor networks, and had provided important comments and feedback. We would also like to thank Nicolas Gillis, John Greer and Todd Wittman for helpful discussions about endmember detection strategies, and Yuanchang Sun for his advice on BSS problems.

\begin{appendix}
\section*{Proof of Lemma \ref{noisezeroconv}}
\begin{proof}
Since $T_\alpha^\delta$ is a minimizer of $J_\alpha$ we can conclude
\begin{eqnarray}
\|T_\alpha^\delta\|_{1,\infty} &\leq& \|T_\alpha^\delta\|_{1,\infty} + \frac{1}{\alpha} \|X^\delta T_\alpha^\delta - X^\delta \|_F^2 \nonumber \\
&=& \frac{1}{\alpha} J_\alpha(T_\alpha^\delta) \nonumber \\
&\leq& \frac{1}{\alpha} J_\alpha(\hat{T}) \nonumber \\
&=& \|\hat{T}\|_{1,\infty} + \frac{1}{\alpha} \|X^\delta \hat{T} - X^\delta\|_F^2 \nonumber \\
&=& \|\hat{T}\|_{1,\infty} + \frac{1}{\alpha} \|(X + N) \hat{T} - (X + N)\|_F^2 \nonumber \\
&=& \|\hat{T}\|_{1,\infty} + \frac{1}{\alpha} \|\underbrace{(X\hat{T} -X)}_{=0} + (N\hat{T} -N)\|_F^2 \nonumber \\
&\leq& \|\hat{T}\|_{1,\infty} + \frac{1}{\alpha} \|N\|_F^2\|\hat{T}-\text{Id}\|_F^2 \nonumber \\
&=& \|\hat{T}\|_{1,\infty} + \frac{\delta^2}{\alpha}\|\hat{T}-\text{Id}\|_F^2.
\end{eqnarray}
Since $\frac{\delta^2}{\alpha}\rightarrow 0$, $T_\alpha^\delta$ is bounded in the $\| \cdot \|_{1,\infty}$ norm and therefore has a convergent subsequence. Let $T_{\alpha_n}^{\delta_n}$ denote such a convergent subsequence and let $\bar{T}$ be its limit. Because $T_{\alpha_n}^{\delta_n} \geq 0$ we also have $\bar{T} \geq 0$. We can now use the above estimate for showing
\begin{eqnarray}
\label{infnorm}
\|\bar{T} \|_{1,\infty} &=& \lim_n \| T_{\alpha_n}^{\delta_n} \|_{1,\infty} \nonumber \\
												&\leq&  \lim_n \big [ \|\hat{T}\|_{1,\infty} + \frac{\delta_n^2}{\alpha_n}\|\hat{T}- \text{Id}\|_F^2 \big] \nonumber \\
												&=&  \|\hat{T}\|_{1,\infty}.
\end{eqnarray}
Furthermore, we have
\begin{eqnarray}
\label{fidel}
\| X \bar{T} - X \|_F^2 &=& \lim_n \| X^{\delta_n}  T_{\alpha_n}^{\delta_n} - X^{\delta_n}  \|_F^2 \nonumber \\
 												&\leq& \lim_n J_{\alpha_n}(T_{\alpha_n}^{\delta_n}) \nonumber \\
 												&\leq& \lim_n J_{\alpha_n}(\hat{T}) \nonumber \\
 											&\leq& \lim_n \|X^{\delta_n} \hat{T} - X^{\delta_n}  \|_F^2 + \alpha_n \|\hat{T}\|_{1,\infty} \nonumber \\
 											&\leq& \lim_n \delta_n^2 \|\hat{T}- \text{Id}\|_F^2 + \alpha_n \|\hat{T}\|_{1,\infty} \nonumber \\
 											&=& 0.
\end{eqnarray}
Now, by the above estimate (\ref{fidel}) we know that $X \bar{T} = X$. Furthermore, by the estimate (\ref{infnorm}) and taking into account that  $\hat{T}$ was a non-negative $\| \cdot \|_{1,\infty}$-minimum norm solution of $XT=X$, we have shown that the limit of our convergent subsequence is also a non-negative $\| \cdot \|_{1,\infty}$-minimum norm solution of $XT=X$.
\end{proof}
\end{appendix}

\bibliographystyle{IEEEbib}
\bibliography{strings,refs}

\end{document}